# The Shattered Gradients Problem: If resnets are the answer, then what is the question?

David Balduzzi [1] Marcus Frean [1] Lennox Leary [1] JP Lewis [1 2] Kurt Wan-Duo Ma [1] Brian McWilliams [3]

## Abstract

A long-standing obstacle to progress in deep learning is the problem of vanishing and exploding gradients. Although, the problem has largely been overcome via carefully constructed initializations and batch normalization, architectures incorporating skip-connections such as highway and resnets perform much better than standard feedforward architectures despite well-chosen initialization and batch normalization. In this paper, we identify the shattered gradients problem. Specifically, we show that the correlation between gradients in standard feedforward networks decays exponentially with depth resulting in gradients that resemble white noise whereas, in contrast, the gradients in architectures with skip-connections are far more resistant to shattering, decaying sublinearly. Detailed empirical evidence is presented in support of the analysis, on both fully-connected networks and convnets. Finally, we present a new "looks linear" (LL) initialization that prevents shattering, with preliminary experiments showing the new initialization allows to train very deep networks without the addition of skip-connections.

## 1. Introduction

Deep neural networks have achieved outstanding performance (Krizhevsky et al., 2012; Szegedy et al., 2015; He et al., 2016b). Reducing the tendency of gradients to vanish or explode with depth (Hochreiter, 1991; Bengio et al., 1994) has been essential to this progress.

Combining careful initialization (Glorot & Bengio, 2010; He et al., 2015) with batch normalization (Ioffe & Szegedy, 2015) bakes two solutions to the vanishing/exploding gradient problem into a single architecture. The He initialization ensures variance is preserved across rectifier layers, and batch normalization ensures that backpropagation through layers is unaffected by the scale of the weights (Ioffe & Szegedy, 2015).

It is perhaps surprising then that residual networks (resnets) *still* perform so much better than standard architectures when networks are sufficiently deep (He et al., 2016a;b). This raises the question: **If resnets are the solution, then what is the problem?** We identify the shattered gradient problem: a previously unnoticed difficulty with gradients in deep rectifier networks that is orthogonal to vanishing and exploding gradients. The shattering gradients problem is that, as depth increases, gradients in standard feedforward networks increasingly resemble white noise. Resnets dramatically reduce the tendency of gradients to shatter.

Our analysis applies at initialization. Shattering should decrease during training. Understanding how shattering affects training is an important open problem.

**Terminology.** We refer to networks without skip connections as feedforward nets—in contrast to residual nets (resnets) and highway nets. We distinguish between the real-valued *output* of a rectifier and its binary *activation*: the activation is 1 if the output is positive and 0 otherwise.

### 1.1. The Shattered Gradients Problem

The first step is to simply look at the gradients of neural networks. Gradients are averaged over minibatches, depend on both the loss and the random sample from the data, and are extremely high-dimensional, which introduces multiple confounding factors and makes visualization difficult (but see section 4). We therefore construct a minimal model designed to eliminate these confounding factors. The minimal model is a neural network $f_{\mathbf{W}} : \mathbb{R} \to \mathbb{R}$ taking scalars to scalars; each hidden layer contains $N = 200$ rectifier neurons. The model is not intended to be applied to real data. Rather, it is a laboratory where gradients can be isolated and investigated.

We are interested in how the gradient varies, at initializa-

*Equal contribution [1]Victoria University of Wellington, New Zealand [2]SEED, Electronic Arts [3]Disney Research, Zürich, Switzerland. Correspondence to: David Balduzzi <dbalduzzi@gmail.com>, Brian McWilliams <brian@disneyresearch.com>.

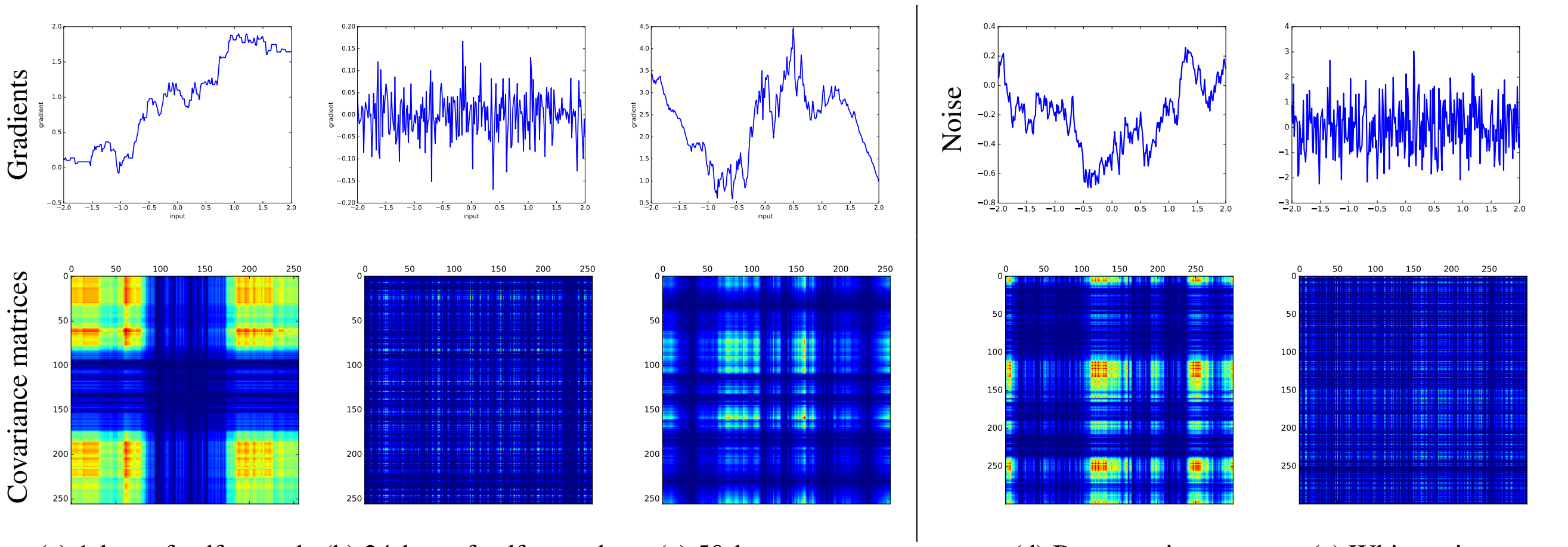

(a) 1-layer feedforward. (b) 24-layer feedforward. (c) 50-layer resnet. (d) Brown noise. (e) White noise.

Figure 1: **Comparison between noise and gradients of rectifier nets with 200 neurons per hidden layer.** *Columns d–e:* brown and white noise. *Columns a–c:* Gradients of neural nets plotted for inputs taken from a uniform grid. The 24-layer net uses mean-centering. The 50-layer net uses batch normalization with $\beta = 0.1$, see Eq. (2).

tion, as a function of the input:

$$\frac{df_{\mathbf{W}}}{dx}(x^{(i)}) \quad \text{where } x^{(i)} \in [-2, 2] \text{ is in a 1-dim grid of } M = 256 \text{ "data points"}. \tag{1}$$

Updates during training depend on derivatives with respect to weights, not inputs. Our results are relevant because, by the chain rule, $\frac{\partial f_{\mathbf{W}}}{\partial w_{ij}} = \frac{\partial f_{\mathbf{W}}}{\partial n_j} \frac{\partial n_j}{\partial w_{ij}}$. Weight updates thus depend on $\frac{\partial f_{\mathbf{W}}}{\partial n_j}$—i.e. how the output of the network varies with the output of neurons in one layer (which are just inputs to the next layer).

The top row of figure 1 plots $\frac{df_{\mathbf{W}}}{dx}(x^{(i)})$ for each point $x^{(i)}$ in the 1-dim grid. The bottom row shows the (absolute value) of the covariance matrix: $|(\mathbf{g} - \bar{\mathbf{g}})(\mathbf{g} - \bar{\mathbf{g}})^\top|/\sigma^2_{\mathbf{g}}$ where $\mathbf{g}$ is the 256-vector of gradients, $\bar{\mathbf{g}}$ the mean, and $\sigma^2_{\mathbf{g}}$ the variance.

If all the neurons were linear then the gradient would be a horizontal line (i.e. the gradient would be constant as a function of $x$). Rectifiers are not smooth, so the gradients are discontinuous.

**Gradients of shallow networks resemble brown noise.** Suppose the network has a single hidden layer: $f_{\mathbf{w},\mathbf{b}}(x) = \mathbf{w}^\top \rho(x \cdot \mathbf{v} - \mathbf{b})$. Following Glorot & Bengio (2010), weights $\mathbf{w}$ and biases $\mathbf{b}$ are sampled from $\mathcal{N}(0, \sigma^2)$ with $\sigma^2 = \frac{1}{N}$. Set $\mathbf{v} = (1, \ldots, 1)$.

Figure 1a shows the gradient of the network for inputs $x \in [-2, 2]$ and its covariance matrix. Figure 1d shows a discrete approximation to brownian motion: $B^N(t) = \sum_{s=1}^{t} W_s$ where $W_s \sim \mathcal{N}(0, \frac{1}{N})$. The plots are strikingly similar: both clearly exhibit spatial covariance structure. The resemblance is not coincidental: section A1 applies Donsker's theorem to show the gradient converges to brownian motion as $N \to \infty$.

**Gradients of deep networks resemble white noise.** Figure 1b shows the gradient of a 24-layer fully-connected rectifier network. Figure 1e shows white noise given by samples $W_k \sim \mathcal{N}(0, 1)$. Again, the plots are strikingly similar.

Since the inputs lie on a 1-dim grid, it makes sense to compute the autocorrelation function (ACF) of the gradient. Figures 2a and 2d compare this function for feedforward networks of different depth with white and brown noise. The ACF for shallow networks resembles the ACF of brown noise. As the network gets deeper, the ACF quickly comes to resemble that of white noise.

Theorem 1 explains this phenomenon. We show that *correlations between gradients decrease exponentially $\frac{1}{2^L}$ with depth in feedforward rectifier networks.*

**Training is difficult when gradients behave like white noise.** The shattered gradient problem is that the spatial structure of gradients is progressively obliterated as neural nets deepen. The problem is clearly visible when inputs are taken from a one-dimensional grid, but is difficult to observe when inputs are randomly sampled from a high-dimensional dataset.

Shattered gradients undermine the effectiveness of algorithms that assume gradients at nearby points are similar such as momentum-based and accelerated methods (Sutskever et al., 2013; Balduzzi et al., 2017). If $\frac{df_{\mathbf{W}}}{dn_j}$ behaves like white noise, then a neuron's effect on the output of the network (whether increasing weights causes the network to output more or less) becomes extremely unstable

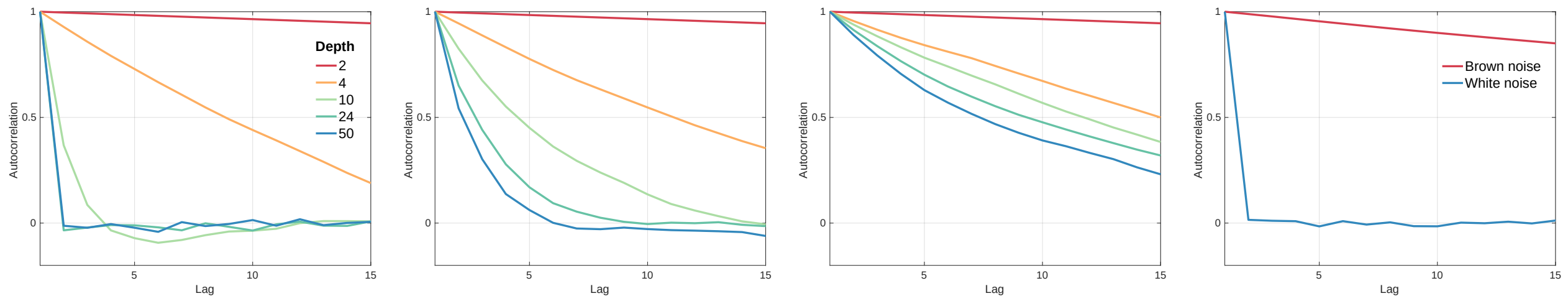

(a) Feedforward nets. (b) Resnets ($\beta = 1.0$). (c) Resnets ($\beta = 0.1$). (d) White and brown noise.

Figure 2: **Autocorrelation Function (ACF).** Comparison of the ACF between white and brown noise, and feedforward and resnets of different depths. Average over 20 runs.

making learning difficult.

**Gradients of deep resnets lie in between brown and white noise.** Introducing skip-connections allows much deeper networks to be trained (Srivastava et al., 2015; He et al., 2016b;a; Greff et al., 2017). Skip-connections significantly change the correlation structure of gradients. Figure 1c shows the concrete example of a 50-layer resnet which has markedly more structure than the equivalent feedforward net (figure 1b). Figure 2b shows the ACF of resnets of different depths. Although the gradients become progressively less structured, they do not whiten to the extent of the gradients in standard feedforward networks—there are still correlations in the 50-layer resnet whereas in the equivalent feedforward net, the gradients are indistinguishable from white noise. Figure 2c shows the dramatic effect of recently proposed $\beta$-rescaling (Szegedy et al., 2016): the ACF of even the 50 layer network resemble brown-noise.

Theorem 3 shows that correlations between gradients decay sublinearly with depth $\frac{1}{\sqrt{L}}$ for resnets with batch normalization. We also show, corollary 1, that modified highway networks (where the gates are scalars) can achieve a *depth independent correlation structure on gradients*. The analysis explains why skip-connections, combined with suitable rescaling, preserve the structure of gradients.

## 1.2. Outline

Section 2 shows that batch normalization increases neural efficiency. We explore how batch normalization behaves differently in feedforward and resnets, and draw out facts that are relevant to the main results.

The main results are in section 3. They explain why gradients shatter and how skip-connections reduce shattering. The proofs are for a mathematically amenable model: fully-connected rectifier networks with the same number of hidden neurons in each layer. Section 4 presents empirical results which show gradients similarly shatter in convnets for real data. It also shows that shattering causes average gradients over minibatches to decrease with depth (relative to the average variance of gradients).

Finally, section 5 proposes the LL-init ("looks linear initialization") which eliminates shattering. Preliminary experiments show the LL-init allows training of extremely deep networks (~200 layers) *without* skip-connections.

## 1.3. Related work

Carefully initializing neural networks has led to a series of performance breakthroughs dating back (at least) to the unsupervised pretraining in Hinton et al. (2006); Bengio et al. (2006). The insight of Glorot & Bengio (2010) is that controlling the variance of the distributions from which weights are sampled allows to control how layers progressively amplify or dampen the variance of activations and error signals. More recently, He et al. (2015) refined the approach to take rectifiers into account. Rectifiers effectively halve the variance since, at initialization and on average, they are active for half their inputs. Orthogonalizing weight matrices can yield further improvements albeit at a computational cost (Saxe et al., 2014; Mishkin & Matas, 2016). The observation that the norms of weights form a random walk was used by Sussillo & Abbott (2015) to tune the gains of neurons.

In short, it has proven useful to treat weights and gradients as random variables, and carefully examine their effect on the variance of the signals propagated through the network. This paper presents a more detailed analysis that considers *correlations* between gradients at different datapoints.

The closest work to ours is Veit et al. (2016), which shows resnets behave like ensembles of shallow networks. We provide a more detailed analysis of the effect of skip-connections on gradients. A recent paper showed resnets have universal finite-sample expressivity and may lack spurious local optima (Hardt & Ma, 2017) but does not explain why deep feedforward nets are harder to train than resnets. An interesting hypothesis is that skip-connections improve performance by breaking symmetries (Orhan, 2017).

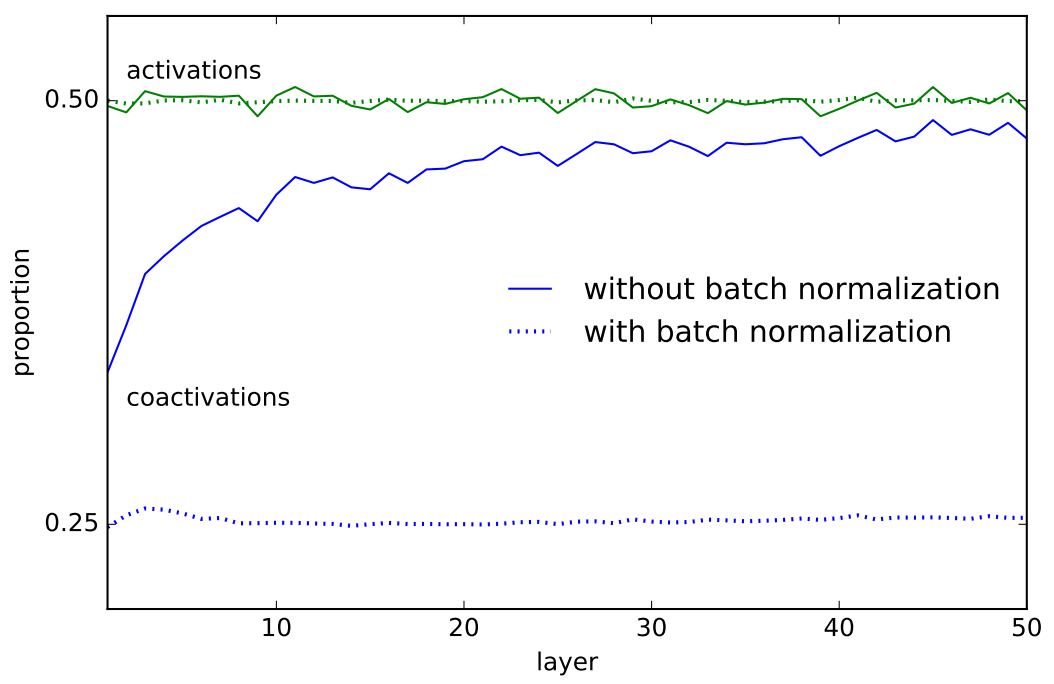

Figure 3: **Activations and coactivations in feedforward networks.** Plots are averaged over 100 fully connected rectifier networks with 100 hidden units per layer. *Without BN:* solid. *With BN:* dotted. *Activations (green):* Proportion of inputs for which neurons in a given layer are active, on average. *Coactivations (blue):* Proportion of distinct pairs of inputs for which neurons are active, on average.

## 2. Observations on batch normalization

Batch normalization was introduced to reduce covariate shift (Ioffe & Szegedy, 2015). However, it has other effects that are less well-known – and directly impact the correlation structure of gradients. We investigate the effect of batch normalization on neuronal activity at initialization (i.e. when it mean-centers and rescales to unit variance).

We first investigate batch normalization's effect on neural activations. Neurons are active for half their inputs on average, figure 3, with or without batch normalization. Figure 3 also shows how often neurons are co-active for two inputs. With batch normalization, neurons are co-active for $\frac{1}{4}$ of distinct pairs of inputs, which is what would happen if activations were decided by unbiased coin flips. Without batch normalization, the co-active proportion climbs with depth, suggesting neuronal responses are increasingly redundant. Resnets with batch normalization behave the same as feedforward nets (not shown).

Figure 4 takes a closer look. It turns out that computing the proportion of inputs causing neurons to be active *on average* is misleading. The distribution becomes increasingly bimodal with depth. In particular, neurons are either always active or always inactive for layer 50 in the feedforward net without batch normalization (blue histogram in figure 4a). Batch normalization causes most neurons to be active for half the inputs, blue histograms in figures 4b,c.

Neurons that are always active may as well be linear. Neurons that are always inactive may as well not exist. It follows that batch normalization increases the *efficiency* with which rectifier nonlinearities are utilized.

The increased efficiency comes at a price. The raster plot for feedforward networks resembles static television noise: the spatial structure is obliterated. Resnets (Figure 4c) exhibit a compromise where neurons are utilized efficiently but the spatial structure is also somewhat preserved. The preservation of spatial structure is quantified via the *contiguity histograms* which counts long runs of consistent activation. Resnets maintain a broad distribution of contiguity even with deep networks whereas batch normalization on feedforward nets shatters these into small sections.

## 3. Analysis

This section analyzes the correlation structure of gradients in neural nets *at initialization*. The main ideas and results are presented; the details provided in section A3.

Perhaps the simplest way to probe the structure of a random process is to measure the first few moments: the mean, variance and covariance. We investigate how the correlation between *typical datapoints* (defined below) changes with network structure and depth. Weaker correlations correspond to whiter gradients. The analysis is for fully-connected networks. Extending to convnets involves (significant) additional bookkeeping.

**Proof strategy.** The covariance defines an **inner product** on the vector space of real-valued random variables with mean zero and finite second moment. It was shown in Balduzzi et al. (2015); Balduzzi (2016) that the gradients in neural nets are sums of **path-weights** over active paths, see section A3. The first step is to observe that path-weights are **orthogonal** with respect to the variance inner product. To express gradients as linear combinations of path-weights is thus to express them over an orthogonal basis.

Working in the path-weight basis reduces computing the covariance between gradients at different datapoints to counting the number of co-active paths through the network. The second step is to count co-active paths and adjust for rescaling factors (e.g. due to batch normalization).

The following assumption is crucial to the analysis:

**Assumption 1** (typical datapoints)**.** *We say* $\mathbf{x}^{(i)}$ *and* $\mathbf{x}^{(j)}$ *are* ***typical datapoints*** *if half of neurons per layer are active for each and a quarter per layer are co-active for both. We assume all pairs of datapoints are typical.*

The assumption will not hold for every pair of datapoints. Figure 3 shows the assumption holds, on average, under batch normalization for both activations and coactivations. The initialization in He et al. (2015) assumes datapoints activate half the neurons per layer. The assumption on co-activations is implied by (and so weaker than) the assumption in Choromanska et al. (2015) that activations are Bernoulli random variables independent of the inputs.

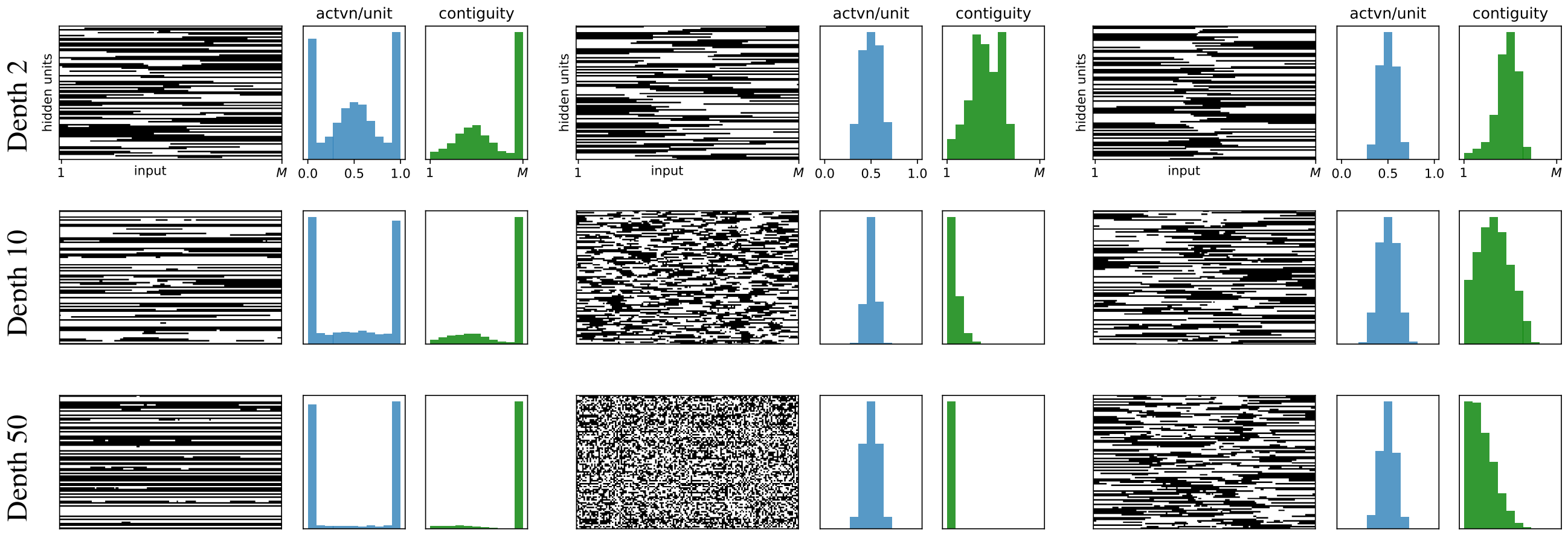

(a) Feedforward net without batch norm. (b) Feedforward with batch normalization. (c) Resnet with batch normalization.

Figure 4: **Activation of rectifiers in deep networks.** *Raster-plots:* Activations of hidden units ($y$-axis) for inputs indexed by the $x$-axis. *Left histogram (activation per unit):* distribution of average activation levels per neuron. *Right histogram (contiguity):* distribution of "contiguity" (length of contiguous sequences of 0 or 1) along rows in the raster plot.

**Correlations between gradients.** Weight updates in a neural network are proportional to

$$\Delta w_{jk} \propto \sum_{i=1}^{\#mb} \sum_{p=1}^{P} \frac{\partial \ell}{\partial f_p} \frac{\partial f_p}{\partial n_k} \frac{\partial n_k}{\partial w_{jk}} (\mathbf{x}^{(i)}).$$

where $f_p$ is the $p^{\text{th}}$ coordinate of the output of the network and $n_k$ is the output of the $k^{\text{th}}$ neuron. The derivatives $\frac{\partial \ell}{\partial f_p}$ and $\frac{\partial n_k}{\partial w_{jk}}$ do not depend on the network's internal structure. We are interested in the middle term $\frac{\partial f_p}{\partial n_k}$, which does. It is mathematically convenient to work with the sum $\sum_{p=1}^{P} f_p$ over output coordinates of the network. Section 4 shows that our results hold for convnets on real-data with the cross-entropy loss. See also remark A2.

**Definition 1.** *Let $\nabla_i := \sum_{p=1}^{P} \frac{\partial f_p}{\partial n}(\mathbf{x}^{(i)})$ be the derivative with respect to neuron $n$ given input $\mathbf{x}^{(i)} \in \mathcal{D}$. For each input $\mathbf{x}^{(i)}$, the derivative $\nabla_i$ is a real-valued random variable. It has mean zero since weights are sampled from distributions with mean zero. Denote the covariance and correlation of gradients by*

$$\mathcal{C}(i,j) = \mathbb{E}[\nabla_i \nabla_j] \text{ and } \mathcal{R}(i,j) = \frac{\mathbb{E}[\nabla_i \nabla_j]}{\sqrt{\mathbb{E}[\nabla_i^2] \cdot \mathbb{E}[\nabla_j^2]}},$$

*where the expectations are w.r.t the distribution on weights.*

### 3.1. Feedforward networks

Without loss of generality, pick a neuron $n$ separated from the output by $L$ layers. The first major result is

**Theorem 1** (covariance of gradients in feedforward nets)**.** *Suppose weights are initialized with variance $\sigma^2 = \frac{2}{N}$ following He et al. (2015). Then*

*a) The variance of the gradient at $\mathbf{x}^{(i)}$ is $\mathcal{C}^{fnn}(i) = 1$.*

*b) The covariance is $\mathcal{C}^{fnn}(i,j) = \frac{1}{2^L}$.*

Part (a) recovers the observation in He et al. (2015) that setting $\sigma^2 = \frac{2}{N}$ preserves the variance across layers in rectifier networks. Part (b) is new. It explains the empirical observation, figure 2a, that gradients in feedforward nets whiten with depth. Intuitively, gradients whiten because the number of paths through the network grows exponentially faster with depth than the fraction of co-active paths, see section A3 for details.

### 3.2. Residual networks

The residual modules introduced in He et al. (2016a) are

$$\mathbf{x}_l = \mathbf{x}_{l-1} + \mathbf{W}^l \rho_{BN}\Big(\mathbf{V}^l \rho_{BN}(\mathbf{x}_{l-1})\Big)$$

where $\rho_{BN}(\mathbf{a}) = \rho(BN(\mathbf{a}))$ and $\rho(a) = \max(0, a)$ is the rectifier. We analyse the stripped-down variant

$$\mathbf{x}_l = \alpha \cdot \big(\mathbf{x}_{l-1} + \beta \cdot \mathbf{W}^l \rho_{BN}(\mathbf{x}_{l-1})\big) \tag{2}$$

where $\alpha$ and $\beta$ are rescaling factors. Dropping $\mathbf{V}^l \rho_{BN}$ makes no essential difference to the analysis. The $\beta$-rescaling was introduced in Szegedy et al. (2016) where it was observed setting $\beta \in [0.1, 0.3]$ reduces instability. We include $\alpha$ for reasons of symmetry.

**Theorem 2** (covariance of gradients in resnets)**.** *Consider a resnet **with batch normalization disabled** and $\alpha = \beta = 1$. Suppose $\sigma^2 = \frac{2}{N}$ as above. Then*

*a) The variance of the gradient at $\mathbf{x}^{(i)}$ is $\mathcal{C}^{res}(i) = 2^L$.*

*b) The covariance is* $\mathcal{C}^{res}(i,j) = \left(\frac{3}{2}\right)^L$.

*The correlation is* $\mathcal{R}^{res}(i,j) = \left(\frac{3}{4}\right)^L$.

The theorem implies there are two problems in resnets without batch normalization: (i) the variance of gradients grows and (ii) their correlation decays exponentially with depth. Both problems are visible empirically.

### 3.3. Rescaling in Resnets

A solution to the exploding variance of resnets is to rescale layers by $\alpha = \frac{1}{\sqrt{2}}$ which yields

$$\mathcal{C}^{\text{res}}_{\alpha=\sqrt{2}}(i) = 1 \text{ and } \mathcal{R}^{\text{res}}_{\alpha=\sqrt{2}}(i,j) = \left(\frac{3}{4}\right)^L$$

and so controls the variance but the correlation between gradients still decays exponentially with depth. Both theoretical predictions hold empirically.

In practice, $\alpha$-rescaling is not used. Instead, activations are rescaled by batch normalization (Ioffe & Szegedy, 2015) and, more recently, setting $\beta \in [0.1, 0.3]$ per Szegedy et al. (2016). The effect is dramatic:

**Theorem 3** (covariance of gradients in resnets with BN and rescaling)**.** *Under the assumptions above, for resnets with batch normalization and $\beta$-rescaling,*

*a) the variance is* $\mathcal{C}^{res}_{\beta,BN}(i) = \beta^2(L-1)+1$;

*b) the covariance*[1] *is* $\mathcal{C}^{res}_{\beta,BN}(i,j) \sim \beta\sqrt{L}$; *and*

*the correlation is* $\mathcal{R}^{res}_{\beta,BN}(i,j) \sim \frac{1}{\beta\sqrt{L}}$.

The theorem explains the empirical observation, figure 2a, that gradients in resnets whiten much more slowly with depth than feedforward nets. It also explains why setting $\beta$ near zero further reduces whitening.

Batch normalization changes the decay of the correlations from $\frac{1}{2^L}$ to $\frac{1}{\sqrt{L}}$. Intuitively, the reason is that the variance of the outputs of layers grows linearly, so batch normalization rescales them by different amounts. Rescaling by $\beta$ introduces a constant factor. Concretely, the model predicts using batch normalization with $\beta = 0.1$ on a 100-layer resnet gives typical correlation $\mathcal{R}^{\text{res}}_{0.1,\text{BN}}(i,j) = 0.7$. Setting $\beta = 1.0$ gives $\mathcal{R}^{\text{res}}_{1.0,\text{BN}}(i,j) = 0.1$. By contrast, a 100-layer feedforward net has correlation indistinguishable from zero.

### 3.4. Highway networks

Highway networks can be thought of as a generalization of resnets, that were in fact introduced slightly earlier (Srivastava et al., 2015; Greff et al., 2017). The standard highway network has layers of the form

$$\mathbf{x}_l = \big(1 - T(\mathbf{x}_{l-1})\big)\cdot \mathbf{x}_{l-1} + T(\mathbf{x}_{l-1})\cdot H(\mathbf{x}_{l-1})$$

where $T(\cdot)$ and $H(\cdot)$ are learned gates and features respectively. Consider the following modification where $\gamma_1$ and $\gamma_2$ are scalars satisfying $\gamma_1^2 + \gamma_2^2 = 1$:

$$\mathbf{x}_l = \gamma_1 \cdot \mathbf{x}_{l-1} + \gamma_2 \cdot \mathbf{W}^l \rho(\mathbf{x}_{l-1})$$

The module can be recovered by judiciously choosing $\alpha$ and $\beta$ in equation (2). However, it is worth studying in its own right:

**Corollary 1** (covariance of gradients in highway networks)**.** *Under the assumptions above, for modified highway networks with $\gamma$-rescaling,*

*a) the variance of gradients is* $\mathcal{C}^{HN}_{\gamma}(i) = 1$; *and*

*b) the correlation is* $\mathcal{R}^{HN}_{\gamma}(i,j) = \left(\gamma_1^2 + \frac{1}{2}\gamma_2^2\right)^L$.

In particular, if $\gamma_1 = \sqrt{1 - \frac{1}{L}}$ and $\gamma_2 = \sqrt{\frac{1}{L}}$ then the correlation between gradients does not decay with depth

$$\lim_{L\to\infty} \mathcal{R}^{\text{HN}}_{\gamma}(i,j) = \frac{1}{\sqrt{e}}.$$

The tradeoff is that the contributions of the layers becomes increasingly trivial (i.e. close to the identity) as $L \to \infty$.

## 4. Gradients shatter in convnets

In this section we provide empirical evidence that the main results also hold for deep convnets using the CIFAR-10 dataset. We instantiate feedforward and resnets with 2, 4, 10, 24 and 50 layers of equivalent size. Using a slight modification of the "bottleneck" architecture in He et al. (2016a), we introduce one skip-connection for every two convolutional layers and both network architectures use batch normalization.

Figures 5a and b compare the covariance of gradients in the first layer of feedforward and resnets ($\beta = 0.1$) with a minibatch of 256 random samples from CIFAR-10 for networks of depth 2 and 50. To highlight the spatial structure of the gradients, the indices of the minibatches were reordered according to a $k$-means clustering ($k = 10$) applied to the gradients of the two-layer networks. The same permutation is used for all networks within a row. The spatial structure is visible in both two-layer networks, although it is more apparent in the resnet. In the feedforward network the structure quickly disappears with depth. In the resnet, the structure remains apparent at 50 layers.

[1]See section A3.4 for exact computations.

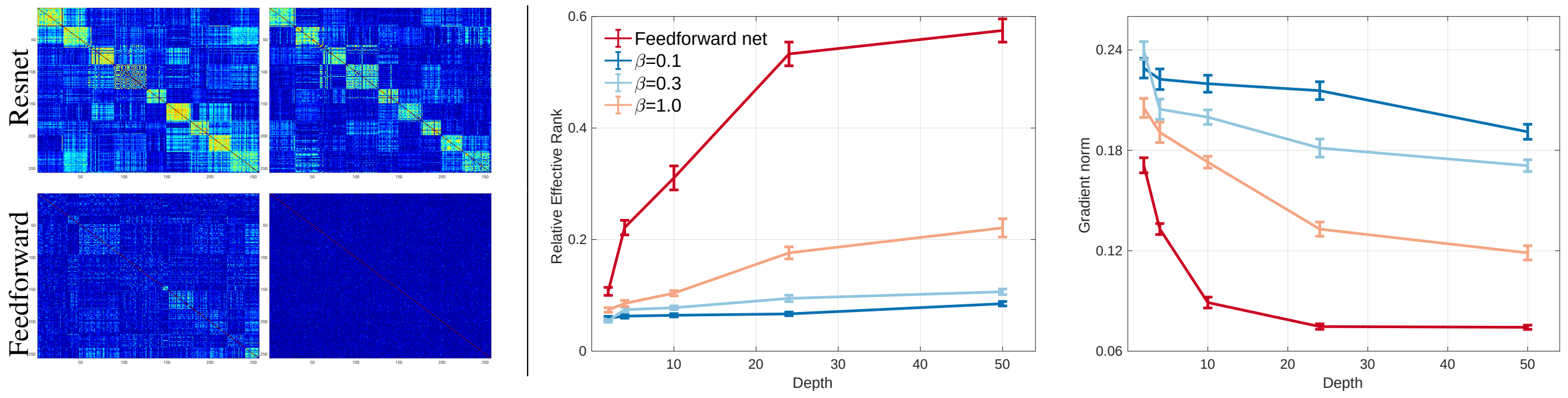

(a) Depth = 2 (b) Depth = 50 (c) Relative effective rank. (d) Gradient norm.

Figure 5: **Results on CIFAR-10**. Figures a–b show the covariance matrices for a single minibatch for feedforward- and resnets. Figures c–d show the relative effective rank and average norms of the gradients averaged over 30 minibatches.

To quantify this effect we consider the "whiteness" of the gradient using *relative effective rank*. Let $\Delta$ be the matrix whose columns are the gradients with respect to the input, for each datapoint $\mathbf{x}^{(i)}$ in a minibatch. The effective rank is $r(\Delta) = \text{tr}(\Delta^\top \Delta)/\|\Delta\|_2^2$ and measures the intrinsic dimension of a matrix (Vershynin, 2012). It is bounded above by the rank of $\Delta$—a matrix with highly correlated columns and therefore more structure will have a lower effective rank. We are interested in the effective rank of the covariance matrix of the gradients relative to a "white" matrix $\mathbf{Y}$ of the same dimensions with i.i.d. Gaussian entries. The *relative* effective rank $r(\Delta)/r(\mathbf{Y})$ measures the similarity between the second moments of $\Delta$ and $\mathbf{Y}$.

Figure 5c shows that the relative effective rank (averaged over 30 minibatches) grows much faster as a function of depth for networks without skip-connections. For resnets, the parameter $\beta$ slows down the rate of growth of the effective rank as predicted by theorem 3.

Figure 5d shows the average $\ell_2$-norm of the gradient in each coordinate (normalized by the standard deviation computed per minibatch). We observe that this quantity decays much more rapidly as a function of depth for feedforward networks. This is due to the effect of averaging increasingly whitening gradients within each minibatch. In other words, the noise within minibatches overwhelms the signal. The phenomenon is much less pronounced in resnets.

Taken together these results confirm the results in section 3 for networks with convolutional layers and show that the gradients in resnets are indeed more structured than those in feedforward nets and therefore do not vanish when averaged within a minibatch. This phenomena allows for the training of very deep resnets.

## 5. The "looks linear" initialization

Shattering gradients are not a problem for linear networks, see remark after equation (1). Unfortunately, linear networks are not useful since they lack expressivity.

The LL-init combines the best of linear and rectifier nets by initializing rectifiers to *look linear*. Several implementations are possible; see Zagoruyko & Komodakis (2017) for related architectures yielding good empirical results. We use concatenated rectifiers or CReLUs (Shang et al., 2016):

$$\mathbf{x} \mapsto \begin{pmatrix} \rho(\mathbf{x}) \\ \rho(-\mathbf{x}) \end{pmatrix}$$

The key observation is that initializing weights with a *mirrored block structure* yields linear outputs

$$\begin{pmatrix} \mathbf{W} & -\mathbf{W} \end{pmatrix} \cdot \begin{pmatrix} \rho(\mathbf{x}) \\ \rho(-\mathbf{x}) \end{pmatrix} = \mathbf{W}\rho(\mathbf{x}) - \mathbf{W}\rho(-\mathbf{x}) = \mathbf{W}\mathbf{x}.$$

The output will cease to be linear as soon as weight updates cause the two blocks to diverge.

An alternative architecture is based on the PReLU introduced in He et al. (2015):

$$\text{PReLU:} \quad \rho_p(x) = \begin{cases} x & \text{if } x > 0 \\ ax & \text{else.} \end{cases}$$

Setting $a = 1$ at initialization obtains a different kind of LL-init. Preliminary experiments, not shown, suggest that the LL-init is more effective on the CReLU-based architecture than PReLU. The reason is unclear.

**Orthogonal convolutions.** A detailed analysis of learning in linear neural networks by Saxe et al. (2014) showed, theoretically and experimentally, that arbitrarily deep linear networks can be trained when initialized with orthogonal weights. Motivated by these results, we use the LL-init in conjunction with orthogonal weights.

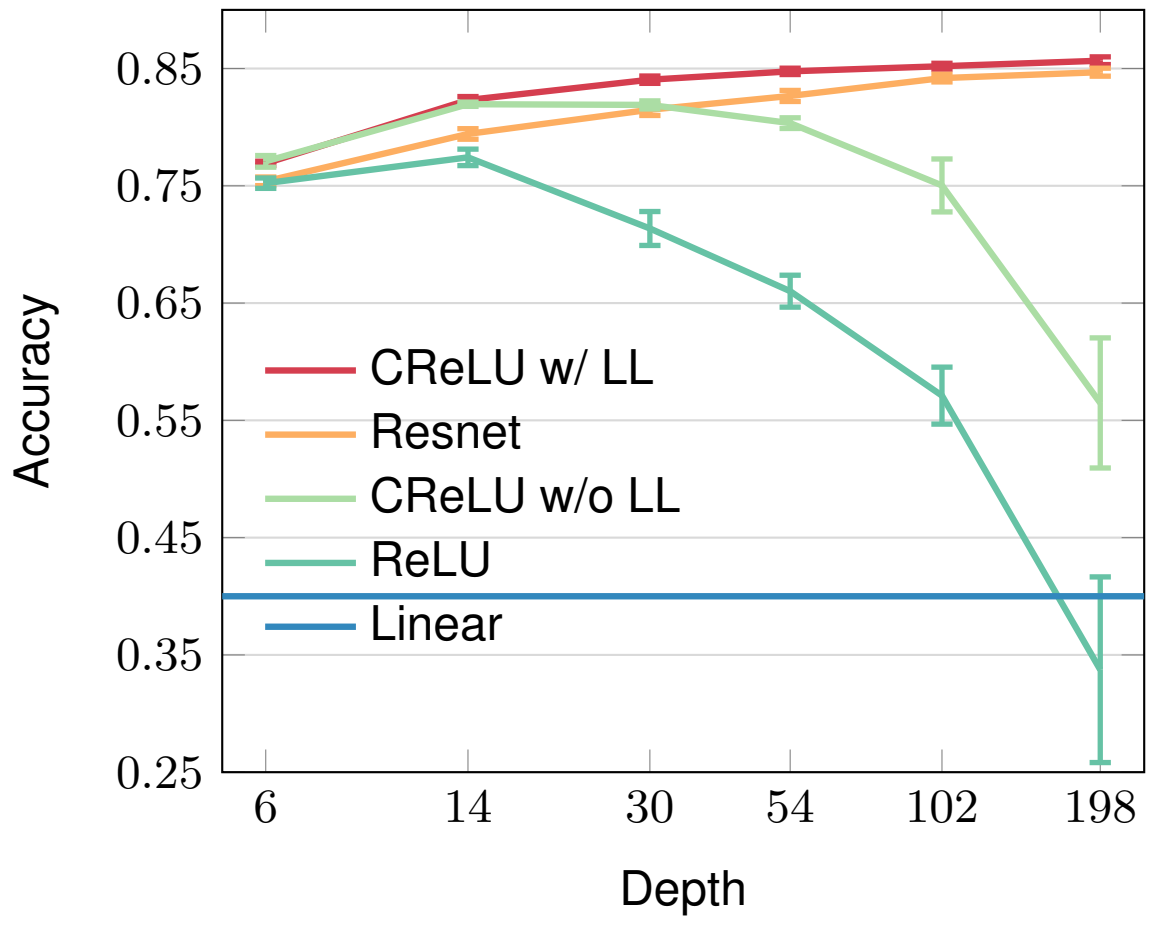

Figure 6: **CIFAR-10 test accuracy**. Comparison of test accuracy between networks of different depths with and without LL initialization.

We briefly describe how we orthogonally initialize a kernel $\mathcal{K}$ of size $A \times B \times 3 \times 3$ where $A \geq B$. First, set all the entries of $\mathcal{K}$ to zero. Second, sample a random matrix $\mathbf{W}$ of size $(A \times B)$ with orthonormal columns. Finally, set $\mathcal{K}[:, :, 2, 2] := \mathbf{W}$. The kernel is used in conjunction with strides of one and zero-padding.

### 5.1. Experiments

We investigated the performance of the LL-init on very deep networks, evaluated on CIFAR-10. The aim was not to match the state-of-the-art, but rather to test the hypothesis that shattered gradients adversely affect training in very deep rectifier nets. We therefore designed an experiment where (concatenated) rectifier nets are and are not shattered at initialization. We find that the LL-init allows to train significantly deeper nets, which confirms the hypothesis.

We compared a CReLU architecture with an orthogonal LL-init against an equivalent CReLU network, resnet, and a standard feedforward ReLU network. The other networks were initialized according to He et al. (2015). The architectures are thin with the number of filters per layer in the ReLU networks ranging from 8 at the input layer to 64, see section A4. Doubling with each spatial extent reduction. The thinness of the architecture makes it particularly difficult for gradients to propagate at high depth. The reduction is performed by convolutional layers with strides of 2, and following the last reduction the representation is passed to a fully connected layer with 10 neurons for classification. The numbers of filters per layer of the CReLU models were adjusted by a factor of $1/\sqrt{2}$ to achieve parameter parity with the ReLU models. The Resnet version of the model is the same as the basic ReLU model with skip-connections after every two modules following He et al. (2016a).

Updates were performed with Adam (Kingma & Ba, 2015). Training schedules were automatically determined by an auto-scheduler that measures how quickly the loss on the training set has been decreasing over the last ten epochs, and drops the learning rate if a threshold remains crossed for five measurements in a row. Standard data augmentation was performed; translating up to 4 pixels in any direction and flipping horizontally with $p = 0.5$.

Results are shown in figure 6. Each point is the mean of 10 trained models. The ReLU and CReLU nets performed steadily worse with depth; the ReLU net performing worse than the linear baseline of 40% at the maximum depth of 198. The feedforward net with LL-init performs comparably to a resnet, suggesting that shattered gradients are a large part of the problem in training very deep networks.

## 6. Conclusion

The representational power of rectifier networks depends on the number of linear regions into which it splits the input space. It was shown in Montufar et al. (2014) that the number of linear regions can grow exponentially with depth (but only polynomially with width). Hence deep neural networks are capable of far richer mappings than shallow ones (Telgarsky, 2016). An underappreciated consequence of the exponential growth in linear regions is the proliferation of discontinuities in the gradients of rectifier nets.

This paper has identified and analyzed a previously unnoticed problem with gradients in deep networks: in a randomly initialized network, the gradients of deeper layers are increasingly uncorrelated. Shattered gradients play havoc with the optimization methods currently in use[2] and may explain the difficulty in training deep feedforward networks even when effective initialization and batch normalization are employed. Averaging gradients over minibatches becomes analogous to integrating over white noise – there is no clear trend that can be summarized in a single average direction. Shattered gradients can also introduce numerical instabilities, since small differences in the input can lead to large differences in gradients.

Skip-connections in combination with suitable rescaling reduce shattering. Specifically, we show that the rate at which correlations between gradients decays changes from exponential for feedforward architectures to sublinear for resnets. The analysis uncovers a surprising and (to us at least) unexpected side-effect of batch normalization. An alternate solution to the shattering gradient problem is to design initializations that do not shatter such as the LL-init. An interesting future direction is to investigate hybrid architectures combining the LL-init with skip connections.

[2] Note that even the choice of a step size in SGD typically reflects an assumption about the correlation scale of the gradients.

# APPENDIX

## A1. Backprop and Brownian Motion

Brownian motion is a stochastic process $\{B_t : t \geq 0\}$ such that

- $B_0 = 0$
- $(B_{t_2} - B_{t_1}) \sim \mathcal{N}(0, t_2 - t_1)$ for any $0 \leq t_1 < t_2$.

- $(B_{t_2} - B_{t_1})$ and $(B_{t_4} - B_{t_3})$ are independent for any $0 \le t_1 < t_2 \le t_3 < t_4$.
- the sample function $t \mapsto B_t(\omega)$ is continuous for almost all $\omega$.

Some important properties of Brownian motion are that

- $B_t \sim \mathcal{N}(0, t)$. In particular $\mathbb{E}[B_t] = 0$ and $\text{Var}[B_t] = t$.
- $\mathbb{E}[B_t B_s] = \min(t, s)$ for any $0 \le s, t$.

The following well known theorem shows how Brownian motion arises as an infinite limit of discrete random walks:

**Theorem** (Donsker). *Let $X_1, \ldots,$ be i.i.d. random variables with mean 0 and variance 1. Let $S_N = \sum_{i=1}^N X_i$. Then the rescaled random walk*

$$B_t^{(N)} = \frac{S_{\lfloor Nt \rfloor}}{\sqrt{N}} \quad \textit{for} \quad t \in [0, 1] \tag{A1}$$

*converges weakly $\lim_{n \to \infty} B^{(N)} = B$ to Brownian motion $B_{t \in [0,1]}$ on the interval $[0, 1]$.*

We are now in a position to demonstrate the connection between the gradients and Brownian motion.

**Proposition A1.** *Suppose weights are sampled from a distribution with mean zero and variance $\sigma^2 = \frac{1}{N}$ per Glorot & Bengio (2010). Then the derivative of $f_{\mathbf{W},\mathbf{b}}$, suitably reparametrized, converges weakly to Brownian motion as $N \to \infty$.*

*Proof.* The derivative of the neural net with respect to its input is:

$$\frac{d}{dx} f_{\mathbf{w},\mathbf{b}}(x) = \sum_{x > b_i} w_i. \tag{A2}$$

If we vary $x$, then the (A2) is a random walk that jumps at points sampled from a Gaussian. In contrast, discrete Brownian motion, (A1), jumps at uniformly spaced points in the unit interval.

Relabel the neurons so the biases are ordered

$$b_1 \le b_2 \le \cdots \le b_N$$

without loss of generality. A rectifier is *active* if its output is nonzero. Let $\boldsymbol{A}(x) = \{i : x > b_i\}$ denote the vector of hidden neurons that are active for input $x$. Ordering the neurons by their bias terms means the derivative only depends on $|\boldsymbol{A}(x)|$, the *number* of active neurons:

$$\frac{d}{dx} f_{\mathbf{w},\mathbf{b}}(x) = \sum_{i=1}^{|\boldsymbol{A}(x)|} w_i.$$

Finally, we can write the derivative as a function of the fraction $t \in [0, 1]$ of neurons that are active:

$$\frac{d}{dx} f_{\mathbf{w},\mathbf{b}}(t) = \sum_{i=1}^{\lfloor Nt \rfloor} w_i.$$

The result follows by Donsker's theorem since the weights are sampled from $\mathcal{N}(0, \frac{1}{N})$. □

## A2. The Karhunen-Loeve theorem

Let $\{X_t : t \in [0, 1]\}$ be a stochastic process with $\mathbb{E}[X_t] = 0$ for all $t$. The covariance function is

$$K(s, t) = \text{Cov}(X_s, X_t) = \mathbb{E}[X_s X_t].$$

Define the associated integral operator $T_K : L_2(\mathbb{R}) \to L_2(\mathbb{R})$ as

$$T_K(\phi)(t) = \int_0^1 K(t, s)\phi(t)\, ds$$

If $K(t, s)$ is continuous in $t$ and $s$ then, by Mercer's theorem, the operator $T_K$ has orthonormal basis of eigenvectors $\mathbf{e}_i(t)$ with associated eigenvalues $\lambda_i$.

**Theorem** (Karhunen-Loeve). *Let*

$$F_i = \int_0^1 X_t \mathbf{e}_i(t)\, dt$$

*Then $\mathbb{E}[F_i] = 0$, $\mathbb{E}[F_i F_j] = 0$ for $i \neq j$, Var$[F_i] = \lambda_i$, and*

$$X_t = \sum_{i=1}^{\infty} F_i \mathbf{e}_i(t)$$

*under uniform convergence in the mean with respect to $t$.*

For example, the eigenvectors and eigenfunctions of Brownian motion, with $K(s, t) = \min(s, t)$, are

$$\mathbf{e}_k(t) = \sqrt{2} \sin\left((k - \frac{1}{2})\pi t\right) \quad \text{and} \quad \lambda_k = \frac{1}{(k - \frac{1}{2})^2 \pi^2}.$$

## A3. Details of the Analysis

**Neural functional analysis.** Functional analysis studies functions and families of functions in vector spaces equipped with a topological structure such as a metric. A fundamental tool is to expand a function in terms of an orthonormal basis $f(x) = \sum_k \alpha_k \mathbf{e}_k(x)$ where the basis satisfies $\langle \mathbf{e}_j(x), \mathbf{e}_k(x) \rangle = \mathbf{1}_{j=k}$. A classical example is the Fourier expansion; a more recent example is wavelets.

A powerful tool for analyzing random processes based on the same philosophy is the Karhunen-Loeve transform. The idea is to represent random processes as linear combinations of orthogonal vectors or functions. For example, principal component analysis is a special case of the Karhunen-Loeve transform.

The weights of a neural network at initialization are random variables. We can therefore model the output of the neural network as a random process indexed by datapoints. That is, for each $\mathbf{x}^{(i)} \in \mathcal{D}$, the output $f_{\mathbf{W}}(\mathbf{x}^{(i)})$ is a random variable. Similarly, the gradients of the neural network form a random process indexed by the data.

The main technical insight underlying the analysis below is that *path-weights* (Balduzzi et al., 2015; Balduzzi, 2016) provide an orthogonal basis relative to the inner product on random variables given by the covariance. The role played by path-weights in the analysis of gradients is thus analogous to the role of sin, cos and exp in Fourier analysis.

### A3.1. Covariance structure of path-sums

Lemma A2 below shows that gradients are sums over products of weights, where the products are "zeroed out" if any neuron along the path is inactive. In this section we develop a minimal mathematical model of path-sums that allows us to compute covariances and correlations between gradients in rectifier networks.

To keep things simple, the model is of a network of $L$ layers each of which contains $N$ neurons. Let $\mathbf{W}$ be a random $(N, N, L-1)$-tensor with entries given by independent random variables with mean zero and variance $\sigma^2$. A **path** is a sequence of numbers $\boldsymbol{\alpha} = (\alpha_1, \ldots \alpha_L) \in [N]^L$. The **path-weight** $\boldsymbol{\alpha}$ is $\mathbf{W}_{\boldsymbol{\alpha}} := \prod_{l=1}^{L-1} \mathbf{W}[\alpha_l, \alpha_{l+1}, l]$, the product of the weights along the path.

Path-weights are random variables. The expected weight of a path is zero. Paths are uncorrelated unless they coincide exactly:

$$\mathbb{E}[\mathbf{W}_{\boldsymbol{\alpha}}] = 0 \text{ and } \mathbb{E}[\mathbf{W}_{\boldsymbol{\alpha}}\mathbf{W}_{\boldsymbol{\beta}}] = \begin{cases} \sigma^{2(L-1)} & \text{if } \boldsymbol{\alpha} = \boldsymbol{\beta} \\ 0 & \text{else.} \end{cases} \tag{A3}$$

**Remark A1.** *Equation (A3) implies that path-weights are orthogonal under the inner product given by covariance.*

An **activation configuration** $\boldsymbol{A}$ is a binary $N \times L$-matrix. Path $\boldsymbol{\alpha}$ is **active** under configuration $\boldsymbol{A}$ if all neurons along the path are active, i.e. if $\boldsymbol{A}_{\boldsymbol{\alpha}} = \prod_{l=1}^{L} \boldsymbol{A}[\alpha_l, l] = 1$, otherwise the path is inactive. The **number of active paths** in configuration $\boldsymbol{A}$ is

$$|\boldsymbol{A}| := \sum_{\boldsymbol{\alpha} \in [N]^L} \boldsymbol{A}_{\boldsymbol{\alpha}}.$$

The **number of co-active paths** in configurations $\boldsymbol{A}$ and $\boldsymbol{B}$ is

$$|\boldsymbol{A} \cap \boldsymbol{B}| := \sum_{\boldsymbol{\alpha} \in [N]^L} \boldsymbol{A}_{\boldsymbol{\alpha}} \cdot \boldsymbol{B}_{\boldsymbol{\alpha}}.$$

Finally, the **path-sum** under configuration $\boldsymbol{A}$ is the sum of the weights of all active paths:

$$\mathrm{p}_{\mathbf{W}}(\boldsymbol{A}) = \sum_{\boldsymbol{\alpha} \in [N]^L} \mathbf{W}_{\boldsymbol{\alpha}} \cdot \boldsymbol{A}_{\boldsymbol{\alpha}}.$$

**Lemma A1.** *Path-sums have mean zero,* $\mathbb{E}[\mathrm{p}_{\mathbf{W}}(\boldsymbol{A})] = 0$, *and covariance*

$$\mathbb{E}[\mathrm{p}_{\mathbf{W}}(\boldsymbol{A}) \cdot \mathrm{p}_{\mathbf{W}}(\boldsymbol{B})] = |\boldsymbol{A} \cap \boldsymbol{B}| \cdot \sigma^{2(L-1)}.$$

*A special case is the variance:*
$\mathbb{E}[\mathrm{p}_{\mathbf{W}}(\boldsymbol{A})^2] = |\boldsymbol{A}| \cdot \sigma^{2(L-1)}$.

*Proof.* The mean is zero since $\mathbb{E}[\mathbf{W}_{\boldsymbol{\alpha}}] = 0$. The cross-terms $\mathbb{E}[\mathbf{W}_{\boldsymbol{\alpha}_1} \cdot \mathbf{W}_{\boldsymbol{\beta}}]$ vanish for $\boldsymbol{\alpha} \neq \boldsymbol{\beta}$ by Eq. (A3), so the covariance simplifies as

$$\begin{aligned} \mathbb{E}[\mathrm{p}_{\mathbf{W}}(\boldsymbol{A})\mathrm{p}_{\mathbf{W}}(\boldsymbol{B})] &= \sum_{\boldsymbol{\alpha},\boldsymbol{\beta} \in [N]^L} \mathbb{E}[\mathbf{W}_{\boldsymbol{\alpha}}\mathbf{W}_{\boldsymbol{\beta}}] \cdot \boldsymbol{A}_{\boldsymbol{\alpha}}\boldsymbol{B}_{\boldsymbol{\beta}} \\ &= \sum_{\boldsymbol{\alpha} \in [N]^L} \mathbb{E}[\mathbf{W}_{\boldsymbol{\alpha}}^2] \cdot \boldsymbol{A}_{\boldsymbol{\alpha}}\boldsymbol{B}_{\boldsymbol{\alpha}} \end{aligned}$$

and the result follows. □

### A3.2. Gradients are path-sums

Consider a network of $L+1$ layers number $0, 1, \ldots L$, where each layer contains $N$ neurons. Let

$$s_L = x_{L,1} + \cdots + x_{L,N}$$

be the sum of the outputs of the neurons in the last layer.

**Lemma A2.** *The derivative*

$$\frac{\partial s_L}{\partial x_{0,i}} = \sum_{\boldsymbol{\alpha} = (i, \alpha_1, \ldots, \alpha_L) \in \{i\} \times [N]^L} \mathbf{W}_{\boldsymbol{\alpha}} \cdot \boldsymbol{A}(\mathbf{x})_{\boldsymbol{\alpha}}.$$

*is the sum of the weights of all active paths from neuron $i$ to the last layer.*

*Proof.* Direct computation. □

**Remark A2.** *The setup of lemma A2 is quite specific. It is chosen for mathematical convenience. In particular, the numerical coincidence that multiplying the number of paths by the variance of the paths yields exactly one, when weights are initialized according to (He et al., 2015), makes the formulas easier on the eye.*

*The theorems are concerned with the large-scale behavior of the variance and covariance as a function of the number of layers (e.g. exponential versus sublinear decay). Their broader implications—but not the precise quantities—are robust to substantial changes to the setup.*

**Proof of theorem 1.**

*Proof.* Lemma A2 implies that gradients are sums over path-weights.

a) By lemma A1 the gradient decomposes as a sum over active paths. There are $N^L$ paths through the network. If half the neurons per layer are active, then there are $|\boldsymbol{A}(\mathbf{x}_i)| = (\frac{N}{2})^L$ active paths. Each path is a product of $L$ weights and so has covariance $\sigma^{2L} = (\frac{2}{N})^L$. Thus, by lemma A1

$$\left(\frac{N}{2}\right)^L \cdot \left(\frac{2}{N}\right)^L = \prod_{l=1}^{L}(1) = 1.$$

b) The number of coactive neurons per layer is $\frac{N}{4}$ and so there are $|\boldsymbol{A}(\mathbf{x}_i) \cap \boldsymbol{A}(\mathbf{x}_j)| = (\frac{N}{4})^L$ coactive paths. By lemma A1 the covariance is

$$\left(\frac{1}{2}\frac{N}{2}\right)^L \cdot \left(\frac{2}{N}\right)^L = \prod_{l=1}^{L}\left(\frac{1}{2}\right) = \frac{1}{2^L}$$

as required. □

### A3.3. Covariance structure of residual path-sums

Derivatives in resnets are path-sums as before. However, skip-connections complicate their structure. We adapt the minimal model in section A3.1 to resnets as follows.

A residual path is a pair $\tilde{\boldsymbol{\alpha}} = (F, \boldsymbol{\alpha})$ where $F \subset [L]$ and $\boldsymbol{\alpha} \in [N]^{|F|}$. The subset $F$ specifies the layers that are *not* skipped; $\boldsymbol{\alpha}$ specifies the neurons in those layers. The length of the path is $l(\tilde{\boldsymbol{\alpha}}) = |F|$. Let $\mathcal{P}^{\text{res}}$ denote the set of all residual paths. Let $F_i$ denote the $i^{\text{th}}$ element of $F$, listed from smallest to largest. Given weight tensor $\mathbf{W}$ as in section A3.1, the weight of path $\tilde{\boldsymbol{\alpha}}$ is

$$\tilde{\mathbf{W}}_{\tilde{\boldsymbol{\alpha}}} = \begin{cases} 1 & \text{if } F = \emptyset \\ \prod_{i=1}^{|F|-1} \mathbf{W}[\alpha_i, \alpha_{i+1}, F_i] & \text{else.} \end{cases}$$

**Remark A3.** *We adopt the convention that products over empty index sets equal one.*

Path $\tilde{\boldsymbol{\alpha}}$ is active under configuration $\boldsymbol{A}$ if $\tilde{\boldsymbol{A}}_{\tilde{\boldsymbol{\alpha}}} = 1$ where

$$\tilde{\boldsymbol{A}}_{\tilde{\boldsymbol{\alpha}}} = \prod_{i=1}^{|F|} \boldsymbol{A}[\alpha_i, F_i]$$

The **residual path-sum** under configuration $\boldsymbol{A}$ is

$$\mathrm{r}_{\mathbf{W}}(\boldsymbol{A}) = \sum_{\tilde{\boldsymbol{\alpha}} \in \mathcal{P}^{\text{res}}} \tilde{\mathbf{W}}_{\tilde{\boldsymbol{\alpha}}} \cdot \tilde{\boldsymbol{A}}_{\tilde{\boldsymbol{\alpha}}}.$$

Restricting to $F = [L]$ recovers the definitions for standard feedforward networks in section A3.1.

The number of co-active paths shared by configurations $\boldsymbol{A}$ and $\boldsymbol{B}$ on the layers in $F$ is

$$|\boldsymbol{A} \cap \boldsymbol{B}|_F = \sum_{\boldsymbol{\alpha} \in [N]^{|F|}} \boldsymbol{A}[\alpha_i, F_i] \cdot \boldsymbol{B}[\alpha_i, F_i].$$

**Lemma A3.** *The covariance between two residual path-sums is*

$$\mathbb{E}[\mathrm{r}_{\mathbf{W}}(\boldsymbol{A}) \cdot \mathrm{r}_{\mathbf{W}}(\boldsymbol{B})] = \sum_{F \subset [L+1]} |\boldsymbol{A} \cap \boldsymbol{B}|_F \cdot \sigma^{2(|F|-1)}$$

*Proof.* Direct computation. □

### A3.4. Residual gradients

**Proof of theorem 2.** The theorem is proved in the setting of lemma A2, see remark A2 for justification.

*Proof.* a) By lemma A3 the variance is

$$\begin{aligned}
\mathbb{E}[\mathrm{r}_{\mathbf{W}}(\boldsymbol{A})^2] &= \sigma^2 \cdot \sum_{F \subset [L]} \left(\frac{N}{2}\right)^{|F|} \sigma^{2(|F|-1)} \\
&= \left(\frac{2}{N}\right) \cdot \sum_{F \subset [L]} \left(\frac{N}{2}\right)^{|F|} \cdot \left(\frac{2}{N}\right)^{|F|-1} \\
&= \sum_{l=0}^{L} \binom{L}{l} = 2^L
\end{aligned}$$

where the final equality follows from the binomial theorem.

b) For the covariance we obtain

$$\begin{aligned}
\mathbb{E}[\mathrm{r}_{\mathbf{W}}(\boldsymbol{A}) \cdot \mathrm{r}_{\mathbf{W}}(\boldsymbol{B})] &= \sigma^2 \cdot \sum_{F \subset [L]} \left(\frac{N}{4}\right)^{|F|} \sigma^{2(|F|-1)} \\
&= \sum_{F \subset [L]} \left(\frac{1}{2}\right)^{|F|} \\
&= \sum_{l=0}^{L} \binom{L}{l} \left(\frac{1}{2}\right)^l \\
&= \left(1 + \frac{1}{2}\right)^L
\end{aligned}$$

by the binomial theorem. □

A convenient way to intuit the computations is to think of each layer as contributing $(1+1)$ to the variance and $(1+\frac{1}{2})$ to the covariance:

$$\mathcal{C}^{\text{res}}(i) = \prod_{l=1}^{L}(1+1) = 2^L \text{ and}$$

$$\mathcal{C}^{\text{res}}(i, j) = \prod_{l=1}^{L}\left(1 + \frac{1}{2}\right) = \left(\frac{3}{2}\right)^L$$

**Proof of theorem 3 when $\beta = 1$.** The theorem is proved in the setting of lemma A2, see remark A2 for justification.

*Proof.* a) Theorem 2 implies that each additional layer (without batch normalization) doubles the contribution to the variance of gradients, which we write schematically as $v_{l+1} = 2v_l = 2^l$, the variance of the $(l+1)^{\text{st}}$ layer is double the $l^{\text{th}}$ layer.

Batch normalization changes the schema to

$$v_{l+1} = v_l + 1$$

where $v_l$ is the variance of the previous layer and $+1$ is added to account for additional variance generated by the non-skip connection (which is renormalized to have unit-variance). The variance of active path sums through $(l+1)$ layers is therefore

$$v_{l+1} = l + 1. \tag{A4}$$

Finally the variance of gradient $\mathcal{C}^{\text{res}}_{BN}(\mathbf{x}_i) = v_L = L$.

b) The above schema for batch normalization can be written

$$v_{l+1} = v_l + \frac{v_l}{l} = v_l\left(1 + \frac{1}{l}\right)$$

where the rescaling factor $\frac{1}{l}$ is the expected variance of the previous layer per Eq. (A4). Unrolling yields

$$v_L = \prod_{l=1}^{L-1}\left(1 + \frac{1}{l}\right) = L.$$

Taking into account the fact that applying batch-normalization to the $l^{\text{th}}$-layer rescales by $\frac{1}{\sqrt{l}}$, the resnet module can be written in expectation as

$$\mathbf{x}_{l+1} = \mathbf{x}_l + \rho_{BN}(\mathbf{W}^{l+1}\mathbf{x}_l) = \mathbf{x}_l + \frac{\rho(\mathbf{W}^{l+1}\mathbf{x}_l)}{\sqrt{l}}.$$

The contribution of each (non-skip) layer to the covariance is half its contribution to the variance since we assume the two inputs are co-active on a quarter of the neurons per layer. The covariance is therefore given by

$$\prod_{l=1}^{L-1}\left(1 + \frac{1}{2}\cdot\frac{1}{l}\right) \sim \sqrt{2L}$$

as required. □

To intuit the approximation, observe that

$$\prod_{l=1}^{L-1}\left(1 + \frac{1}{2l}\right)\cdot\left(1 + \frac{1}{2l-1}\right) = \prod_{l=1}^{2L-2}\left(1 + \frac{1}{l}\right) = 2L-1$$

Since $\left(1 + \frac{1}{2l}\right) \sim \left(1 + \frac{1}{2l-1}\right)$, rewrite as

$$\prod_{l=1}^{L-1}\left(1 + \frac{1}{2l}\right)^2 \sim \prod_{l=1}^{2L-2}\left(1 + \frac{1}{l}\right) = 2L - 1$$

and so $\prod_{l=1}^{L-1}\left(1 + \frac{1}{2l}\right) \sim \sqrt{L}$.

Numerically we find $\prod_{l=1}^{L-1}\left(1 + \frac{1}{2l}\right) \sim \sqrt{\frac{4}{\pi}(L+1)}$ to be a good approximation for large $L$.

**Proof of theorem 3 for general $\beta$.** The theorem is proved in the setting of lemma A2, see remark A2 for justification.

*Proof.* a) The introduction of $\beta$-rescaling changes the schema to

$$v_{l+1} = v_l\left(1 + \frac{\beta^2}{\beta^2(l-1)+1}\right).$$

The proof then follows from the observation that

$$\prod_{l=1}^{L-1}\left(1 + \frac{\beta^2}{\beta^2(l-1)+1}\right) = \beta^2(L-1)+1.$$

b) The covariance is given by

$$\prod_{l=1}^{L-1}\left(1 + \frac{1}{2}\frac{\beta^2}{\beta^2(l-1)+1}\right) \sim \beta\sqrt{L}$$

by similar working to when $\beta = 1$. □

### A3.5. Highway gradients

**Proof of corollary 1.** The theorem is proved in the setting of lemma A2, see remark A2 for justification.

*Proof.* The variance is given by

$$\prod_{l=1}^{L}\left(\gamma_1^2 + \gamma_2^2\right) = 1$$

and the covariance by

$$\prod_{l=1}^{L}\left(\gamma_1 + \frac{1}{2}\gamma_2\right) = \left(\gamma_1^2 + \frac{1}{2}\gamma_2^2\right)^L$$

by analogous working to the previous theorems. □

Setting $\gamma_1 = \sqrt{1 - \frac{1}{L}}$ and $\gamma_2 = \sqrt{\frac{1}{L}}$ obtains

$$\begin{aligned}\mathcal{C}^{HN}_{\gamma}(\mathbf{x}_i, \mathbf{x}_j) &= \left(1 - \frac{1}{L} + \frac{1}{2}\frac{1}{L}\right)^L \\ &= \left(1 - \frac{1}{2}\frac{1}{L}\right)^L \\ &\xrightarrow[\infty]{L} e^{-\frac{1}{2}}\end{aligned}$$

by standard properties of the constant $e$.

## A4. Details on architecture for figure 6

| |
|---|
| $r$ modules with 8 filters each |
| Downsampling module with 16 filters<br>$r-1$ modules with 16 filters each |
| Downsampling module with 32 filters<br>$r-1$ modules with 32 filters each |
| Downsampling module with 64 filters<br>$r-1$ modules with 64 filters each |
| Downsampling module with 64 filters |
| Flattening layer |
| FC layer to output (width 10) |